\pdfoutput=1
\documentclass[11pt,a4paper]{article}
\usepackage[dvipsnames,table,xcdraw]{xcolor}
\usepackage[hyperref]{emnlp2020}
\usepackage{times}
\usepackage{latexsym}

\usepackage{xargs}
\usepackage{hyperref}
\usepackage{url}
\usepackage{graphicx,xcolor}
\usepackage{booktabs,subcaption,amsmath}
\usepackage{tikz}
\usepackage{tikz-qtree}
\usepackage[colorinlistoftodos,prependcaption,textsize=tiny]{todonotes}

\usepackage{microtype}

\aclfinalcopy 

\usepackage[top=30mm,bottom=40mm,inner=30mm,outer=35mm]{geometry}
\title{Conversational Semantic Parsing}

\author{Armen Aghajanyan \quad Jean Maillard \quad Akshat Shrivastava \quad Keith Diedrick\\
\textbf{Mike Haeger\quad Haoran Li\quad Yashar Mehdad\quad Ves Stoyanov}\\
\textbf{Anuj Kumar \quad Mike Lewis \quad Sonal Gupta}\\
Facebook \\
\texttt{\{armenag,jeanm,akshats,kdiedrick,mhaeger,aimeeli\}@fb.com} \\
\texttt{\{mehdad,ves,anujk,mikelewis,sonalgupta\}@fb.com} \\
}

\date{}

\begin{document}
\maketitle

\begin{abstract}
The structured representation for semantic parsing in task-oriented assistant systems is geared towards simple understanding of one-turn queries. Due to the limitations of the representation, the session-based properties such as co-reference resolution and context carryover are processed downstream in a pipelined system. In this paper, we propose a semantic representation for such task-oriented conversational systems that can represent concepts such as co-reference and context carryover, enabling comprehensive understanding of queries in a session. We release a new session-based, compositional task-oriented parsing dataset of 20k sessions consisting of 60k utterances. Unlike Dialog State Tracking Challenges, the queries in the dataset have compositional forms. We propose a new family of Seq2Seq models for the session-based parsing above, which achieve better or comparable performance to the current state-of-the-art on ATIS, SNIPS, TOP and DSTC2. Notably, we improve the best known results on DSTC2 by up to 5 points for slot-carryover.

\end{abstract}

\section{Introduction}
 At the core of conversational assistants lies the semantic representation, which provides a structured description of tasks supported by the assistant.  Traditional dialog systems operate through a flat representation, usually composed of a single intent and a list of slots with non-overlapping content from the utterance \citep{bapna2017towards,gupta2018semantic}.  Although flat representations are trivial to model with standard intent/slot tagging models, the semantic representation is fundamentally limiting. \citet{gupta2018semantic} explored the limitations of flat representations and proposed a compositional generalization which allowed slots to contain nested intents while allowing easy modeling through neural shift-reduce parsers such as RNNG \citep{dyer2016recurrent}.

 Our contributions are the following:
 \begin{itemize}
    \item We explore the limitations of this compositional form and propose an extension which overcomes these limitations that we call \emph{decoupled representation}.
    \item To parse this more complicated representation, we propose a family of Seq2Seq models based off the Pointer-Generator architecture that set state of the art in multiple semantic parsing and dialog tasks \citep{see2017get}.
    \item To further advance session based task oriented semantic parsing, we release a publicly available set with 60k utterances constituting roughly 20k sessions.
 \end{itemize}

\section{Semantic Representation}
The compositional extension proposed by \citet{gupta2018semantic} overcame the limitation of classical intent-slot frameworks by allowing nested intents in slots. But to maintain an easily model-able structure the following constraint was introduced: \textit{the in-order traversal of the compositional semantic representation must reconstruct the utterance}. Following this constraint it is possible to use discriminative neural shift reduce parsers such as RNNG to parse into this form \citep{dyer2016recurrent}.

Although at face value this constraint seems reasonable, it has non-trivial implications for both the semantic parsing component (NLU) and downstream components in conversational assistants.

\subsection{Surpassing Utterance Level Limitations with Decoupled Form}

First we'll take a look at the space of utterances that can be covered by the compositional representation. One fundamental problem with the in-order constraint is that it disallows long-distance dependencies within the semantic representation. For example, the utterance \textit{On Monday, set an alarm for 8am.} would optimally have a single date-time slot: \texttt{[SL\_DATETIME 8am on Monday]}. But, because \textit{8am} and \textit{on Monday} are at opposite ends of the utterance, there is no way to construct a semantic parse tree with a single date-time slot. \citet{gupta2018semantic} mentioned this problem, but had some empirical data showing that utterances with long-distance dependencies are rare in English. Although this might be true, having fundamental limitations on what type of utterances can be supported even with a complete ontology is concerning.

In English, discontinuities are restricted in occurrence, despite emerging naturally within certain patterns, because English is a configurational language, which uses strongly marked word order to impart some level of semantic information \citep{chomsky1981lectures}. Beyond English, however, there are numerous world languages that are non-configurational and have much freer or potentially completely free word order. Non-configurational languages may often present the same semantic information through the use of Case Markers, Declensions, or other systems. The relatively free word order this allows creates much less emphasis on the collocation of a semantic unit's tokens. Therefore, as conversational assistants progress toward multiple languages it's important to consider that constraints that are acceptable if only English is considered will not analogously scale to other languages.

A simple solution is to convert a standard compositional intent-slot parse into a logical form containing two label types (slot and intent), with no constraints over intent spans. This is trivially accomplished by removing all text in the compositional semantic parse that does not appear in a leaf slot. We call this form of semantic parse the \emph{decoupled} semantic representation, due to the semantic representation not being tightly coupled with the original utterance.

\begin{figure*}[h]
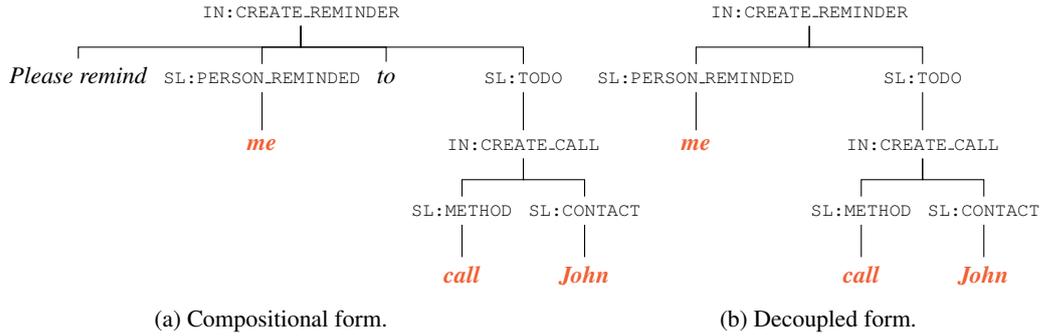

   \centering
   \begin{subfigure}{0.49\textwidth}
      \centering\small
      \tikzset{level distance=25pt,sibling distance=0pt}
      \tikzset{edge from parent/.style={draw, edge from parent path={(\tikzparentnode.south) -- +(0,-8pt) -| (\tikzchildnode)}}}
      \Tree [.\texttt{\scriptsize IN:CREATE\_REMINDER} \emph{Please remind} [.\texttt{\scriptsize SL:PERSON\_REMINDED} \emph{\textcolor{RedOrange}{\textbf{me}}} ] \emph{to} [.\texttt{\scriptsize SL:TODO} [.\texttt{\scriptsize IN:CREATE\_CALL} [.\texttt{\scriptsize SL:METHOD} \emph{\textcolor{RedOrange}{\textbf{call}}} ] [.\texttt{\scriptsize SL:CONTACT} \emph{\textcolor{RedOrange}{\textbf{John}}} ] ] ] ]
      \caption{Compositional form.}\label{subfig:seqlogical-form}
   \end{subfigure}
   \begin{subfigure}{0.49\textwidth}
      \centering\small
      \tikzset{level distance=25pt,sibling distance=0pt}
      \tikzset{edge from parent/.style={draw, edge from parent path={(\tikzparentnode.south) -- +(0,-8pt) -| (\tikzchildnode)}}}
      \Tree [.\texttt{\scriptsize IN:CREATE\_REMINDER} [.\texttt{\scriptsize SL:PERSON\_REMINDED} \emph{\textcolor{RedOrange}{\textbf{me}}} ] [.\texttt{\scriptsize SL:TODO} [.\texttt{\scriptsize IN:CREATE\_CALL} [.\texttt{\scriptsize SL:METHOD} \emph{\textcolor{RedOrange}{\textbf{call}}} ] [.\texttt{\scriptsize SL:CONTACT} \emph{\textcolor{RedOrange}{\textbf{John}}} ] ] ] ]
      \caption{Decoupled form.}\label{subfig:decoupled-form}
   \end{subfigure}
   \caption{Compositional and decoupled semantic representations for the single utterance ``Please remind me to call John''.}\label{fig:rep-single-utt}
\end{figure*}

Figure~\ref{fig:rep-single-utt} shows a side by side example of compositional and decoupled semantic representations for the utterance \textit{Please remind me to call John}.

\subsection{Session Based Limitations}
Because traditional conversational systems historically have had a clear separation between utterance level semantic parsing and dialog systems (which stitch together utterance level information into sessions), semantic representations have not focused on session-based representations. Integrating session information into semantic parsers has been limited to refinement-based approaches.

\begin{figure}
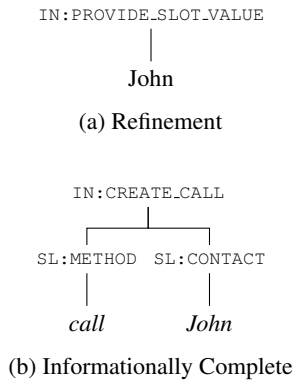

   \centering
   \begin{subfigure}{0.49\textwidth}
      \centering\small
      \tikzset{level distance=25pt,sibling distance=0pt}
      \tikzset{edge from parent/.style={draw, edge from parent path={(\tikzparentnode.south) -- +(0,-8pt) -| (\tikzchildnode)}}}
      \Tree [.\texttt{\scriptsize IN:PROVIDE\_SLOT\_VALUE} John ]
      \caption{Refinement}\label{subfig:refinement}
   \end{subfigure}
   
   \begin{subfigure}{0.49\textwidth}
      \vspace{15pt}
      \centering\small
      \tikzset{level distance=25pt,sibling distance=0pt}
      \tikzset{edge from parent/.style={draw, edge from parent path={(\tikzparentnode.south) -- +(0,-8pt) -| (\tikzchildnode)}}}
      \Tree [.\texttt{\scriptsize IN:CREATE\_CALL} [.\texttt{\scriptsize SL:METHOD} \emph{call} ] [.\texttt{\scriptsize SL:CONTACT} \emph{John} ] ]
      \caption{Informationally Complete}\label{subfig:complete}
   \end{subfigure}
   \caption{\textbf{Refinement} and \textbf{Complete} session based semantic representations for the utterance ``call''.}\label{fig:refinement-vs-complete}
\end{figure}

Figure~\ref{fig:refinement-vs-complete} shows an example of refinement and informationally complete based approaches to semantic parsing.
The refinement approach delegates responsibility of session-based semantic parsing to a separate dialog component. Consequently, refinement approaches tend to have a very limited ontology due to the semantic parser operating over a fixed input (non-session utterances).

Predicting what slot to use for refining works for flat semantic representations, but it is non-trivial to extend to compositional or decoupled. The position of a slot in a flat semantic representation is not meaningful, thus it is sufficient to only predict the slot without specifying its position in the parse. But both compositional and decoupled extensions to intent-slot parsing vary semantically by the position of the slot (or nested intent).

\begin{figure*}[h]
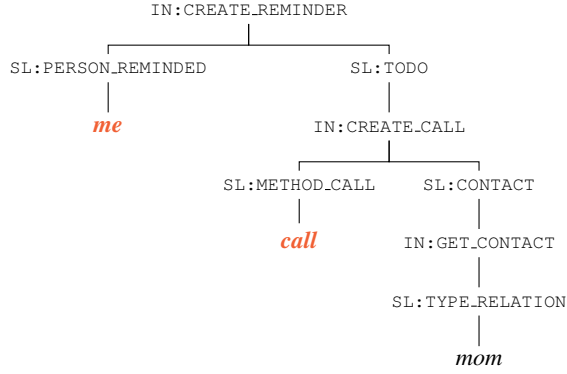

   \centering
   \small
     \centering
     \centering\small
      \tikzset{level distance=22pt,sibling distance=0pt}
      \tikzset{edge from parent/.style={draw, edge from parent path={(\tikzparentnode.south) -- +(0,-8pt) -| (\tikzchildnode)}}}
     \Tree [.\texttt{\scriptsize IN:CREATE\_REMINDER} [.\texttt{\scriptsize SL:PERSON\_REMINDED} \emph{\textcolor{RedOrange}{\textbf{me}}} ] [.\texttt{\scriptsize SL:TODO} [.\texttt{\scriptsize IN:CREATE\_CALL} [.\texttt{\scriptsize SL:METHOD\_CALL} \emph{\textcolor{RedOrange}{\textbf{call}}} ] [.\texttt{\scriptsize SL:CONTACT} [.\texttt{\scriptsize IN:GET\_CONTACT} [.\texttt{\scriptsize SL:TYPE\_RELATION} [.\emph{mom} ] ] ] ] ] ] ]
   \caption{Sample session with complex slot-carryover: ``Is mom available?'' -- ``Remind me to call''}
   \label{fig:complex-slot}
\end{figure*}

We present an example in Figure~\ref{fig:complex-slot}. Given the followup utterance \textit{remind me to call}, a classical system would need to carry over the whole \texttt{CONTACT} slot, but the question is to where? The semantic parse is not flat. The slot could be carried over to the \texttt{CREATE\_REMINDER} intent or the nested \texttt{GET\_CONTACT} intent. So, if we were to extend classical slot carryover, we not only would need to predict what slot to carry over from the conversation, but what intent within the current semantic parse to place it under. We propose a new paradigm that does joint classical semantic parsing with co-reference resolution and slot-carryover.

\subsection{Session Based Semantic Parsing}
We present a simple extension to the decoupled paradigm of intent-slot semantic parsing by introduction of a new reference (\texttt{REF}) label type. The \texttt{REF} label type contains two elements in its set to represent co-references and slot-carryover as separate operations. Co-references can be seen as an explicit reference, namely a reference conditioned on an explicit word, while slot-carryover is treated as an implicit reference (conditioned by relevant contextual information).

\begin{figure*}[h!]
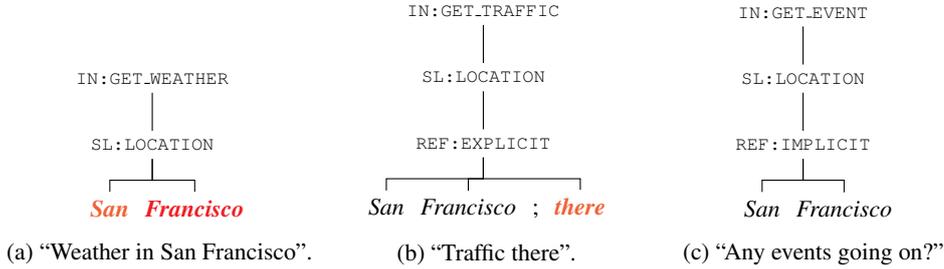

   \centering
   \begin{subfigure}{0.29\textwidth}
      \centering\small
      \vspace{25pt}
      \tikzset{level distance=25pt,sibling distance=0pt}
      \tikzset{edge from parent/.style={draw, edge from parent path={(\tikzparentnode.south) -- +(0,-8pt) -| (\tikzchildnode)}}}
      \Tree [.\texttt{\scriptsize IN:GET\_WEATHER} [.\texttt{\scriptsize SL:LOCATION} \emph{\textcolor{RedOrange}{\textbf{San}}} \emph{\textcolor{Red}{\textbf{Francisco}}} ] ]
      \caption{``Weather in San Francisco''.}\label{subfig:session-1}
   \end{subfigure}
   \begin{subfigure}{0.29\textwidth}
      \centering\small
      \tikzset{level distance=25pt,sibling distance=0pt}
      \tikzset{edge from parent/.style={draw, edge from parent path={(\tikzparentnode.south) -- +(0,-8pt) -| (\tikzchildnode)}}}
      \Tree [.\texttt{\scriptsize IN:GET\_TRAFFIC} [.\texttt{\scriptsize SL:LOCATION} [.\texttt{\scriptsize REF:EXPLICIT} \emph{San} \emph{Francisco} \edge[draw=none] ; {;} \emph{\textcolor{RedOrange}{\textbf{there}}} ] ] ]
      \caption{``Traffic there''.}\label{subfig:session-2}
   \end{subfigure}
   \begin{subfigure}{0.29\textwidth}
      \centering\small
      \tikzset{level distance=25pt,sibling distance=0pt}
      \tikzset{edge from parent/.style={draw, edge from parent path={(\tikzparentnode.south) -- +(0,-8pt) -| (\tikzchildnode)}}}
      \Tree [.\texttt{\scriptsize IN:GET\_EVENT} [.\texttt{\scriptsize SL:LOCATION} [.\texttt{\scriptsize REF:IMPLICIT} \emph{San} \emph{Francisco} ] ] ]
      \caption{``Any events going on?''}\label{subfig:session-3}
   \end{subfigure}
   \caption{Decoupled semantic representations for a three-utterance session.}
   \label{fig:session}
\end{figure*}

As an example, refer to the sample session with decoupled semantic parses in Figure~\ref{fig:session}

\section{Model}

\subsection{Sequence-to-Sequence Architecture}

The decoupled semantic parsing model is an extension of the very common sequence-to-sequence learning approach \citep{seq2seq}, with the \emph{source sequence} being the utterance and the \emph{target sequence} being a linearized version of the target tree. Trees are linearized by bracketing them, using the same approach as \citet{grammar_as_foreign_lang}. The decoupled tree in Fig. \ref{subfig:decoupled-form}, for example, would be linearized to the following target sequence: \texttt{[IN:CREATE\_REMINDER}, \texttt{[SL:PERSON\_REMINDED}, \texttt{me}, \texttt{]}, \texttt{[SL:TODO}, ..., \texttt{]}. After tokenization, an encoder processes the source tokens $w_i$ and produces corresponding encoder hidden states:

$$
   \boldsymbol{e}_1, ..., \boldsymbol{e}_T = \operatorname{Encoder}(w_1, ..., w_T)
$$
where the encoder, in our experiments, is either a standard bidirectional LSTM or a transformer.

In spite of its drawbacks, the rigid structure of the compositional semantic trees (Fig. \ref{subfig:seqlogical-form}) has the advantage of readily mapping to the RNNG formalism and its inductive biases. The decoupled semantic representation, being more flexible, does not have such an easily exploitable form -- but we can still exploit whatever structure exists. The tokens of the linearized decoupled representation (the \emph{target sequence}) can always be divided into two classes: utterance tokens that are already present in the source sequence -- which form the leaves of the tree -- and ontology symbols. Taking again the example tree of Fig. \ref{subfig:decoupled-form}, \emph{me}, \emph{call}, and \emph{John} are all tokens from the utterance, while \texttt{[IN:CREATE\_REMINDER}, \texttt{[SL:PERSON\_REMINDED}, \texttt{]}, etc., are ontology symbols. This partition is reflected in the structure of the decoder: at every decoding step, the model can either generate an element from the ontology, or copy a token from the source sequence via a mechanism analogous to the pointer-generator network of \citet{see2017get}.
At decoding time step $t$, the decoder is fed with the encoder's outputs and produces a vector of features $\boldsymbol{x}_t$, which is used to compute an \emph{ontology generation distribution} $\boldsymbol{p}^\text{g}_t$:
\begin{align*}
\boldsymbol{x}_t &= \operatorname{Decoder}\left(\boldsymbol{e}_1, ..., \boldsymbol{e}_t; \boldsymbol{d}_{t-1}; \boldsymbol{s}_{t-1}\right),\\
\boldsymbol{p}^\text{g}_t &= \operatorname{softmax}\left(\operatorname{Linear}_\text{g}[\boldsymbol{x}_t]\right),
\end{align*}
where $d_{t-1}$ is the previous output of the decoder, $s_{t-1}$ is the decoder's incremental state, and $\operatorname{Linear}_\theta[\boldsymbol{x}]$ is short-hand for an affine transformation with parameters $\theta$, i.e. $\mathbf{W}^\theta \boldsymbol{x} + \boldsymbol{b}^\theta$. The decoder's features are also used to calculate the attention distribution -- using multi-head attention \citep{transformer} -- which then serves to produce the \emph{utterance copy distribution} $\boldsymbol{p}^\text{c}_t$:
\begin{align*}
\boldsymbol{p}^\text{c}_t, \boldsymbol{\omega}_t &= \operatorname{MhAttention}\left(\boldsymbol{e}_1, ..., \boldsymbol{e}_t; \operatorname{Linear}_\text{c}[\boldsymbol{x}_t]  \right),\\
p^\alpha_t &= \operatorname{\sigma}\left( \operatorname{Linear}_\alpha\left[\boldsymbol{x}_t \| \boldsymbol{\omega}_t\right] \right),
\end{align*}
where $\sigma(x) = \frac{1}{1+e^{-x}}$ is the standard sigmoid function, $\|$ indicates concatenation, and $\operatorname{MhAttention}$ indicates multi-head attention which returns, respectively, the attention distribution and its weights. Finally, the extended probability distribution is computed as a mixture of the ontology generation and utterance copy distributions:
$$
\boldsymbol{p}_t = p^\alpha_t \cdot \boldsymbol{p}^\text{g}_t + \left(1-p^\alpha_t\right) \cdot \boldsymbol{p}^\text{c}_t.
$$

\subsection{Encoder and Decoder}

We experiment with two main variants of the decoupled model: one based on recurrent neural networks, and one based on the transformer architecture \citep{transformer}.

\paragraph{RNN} Our base model uses two distinct stacked bidirectional LSTMs as the encoder and stacked unidirectional LSTMs as the decoder. Both consist of two layers of size 512, with randomly initialized embeddings of size 300. The base model is optimized with LAMB while others are optimized with Adam, using parameters $\beta_1=0.9$, $\beta_2=0.999$, $\epsilon=10^{-8}$, and L2 penalty $10^{-5}$ \citep{kingma2014adam}. The learning rate is found separately for each experiment via hyperparameter search. We also use stochastic weight averaging \citep{swa}, and exponential learning rate decay. For an extended version of this model, we also try incorporating contextualized word vectors, by augmenting the input with ELMo embeddings \citep{elmo}.

\paragraph{Transformer} We also experiment with two further variants of the model, that replace encoder and decoder with transformers. In the first variant, the encoder is initialized with RoBERTa \citep{roberta}, a pretrained language model. The decoder is a randomly initialized 3-layer transformer, with hidden size 512 and 4 attention heads. In the second variant, we initialise both encoder and decoder with BART \citep{lewis2019bart}, a sequence-to-sequence pretained model. Both encoder and decoder consist of 12 layers with hidden size 1024.  We train these with stochastic weight averaging \citep{swa}, and determine optimal hypermarameters on the validation sets.

\section{Experiments}
\subsection {Session Based Task Oriented Parsing}
To incentivize further research into session based semantic parsing through the decoupled intent-slot paradigm we are releasing 20 thousand annotated sessions in 4 domains: calling, weather, music and reminder. We also allow for mixtures of domains within a session.

The data was collected in two stages. First we asked crowdsourced workers to write sessions (both from the users perspective as well as the Assistant's output) tied to certain domains. Once we vetted the sessions, we asked a second group of annotators to annotate the user input per session. Each session was given to three separate annotators. We used majority voting to automatically resolve the correct parse when possible. In the cases where there was no agreement, we selected the maximum informative parse which abode by the labeling representations semantic constraints. The annotator agreement rate was 55\%, while our final chosen semantic parses were correct 94\% of the time. The large delta between the two numbers is due to multiple correct semantic parses existing for the same session.

We open source SB-TOP in the following link: \url{http://dl.fbaipublicfiles.com/sbtop/SBTOP.zip}. More information about the dataset can be found in the Table~\ref{table:sbtop_stats} in the Appendix.
\subsection{Semantic Parsing}

We evaluate the decoupled model on five semantic parsing datasets, four public and one internal. All but two are annotated with compositional semantic representations and the other with the standard flat intent-slot representation. In order to apply the decoupled models to them, we follow a mechanical procedure to transform the annotations to decoupled representations: all utterance tokens which are not part of a slot are stripped. This procedure effectively turns the tree of Fig. \ref{subfig:seqlogical-form} into the tree of Fig. \ref{subfig:decoupled-form}. We note that this procedure for all compositional and flat intent-slot data available is reversible, therefore we can convert from decoupled back to source representation.

The first public dataset is TOP \citep{gupta2018semantic}, which consists of over 31k training utterances covering the navigation, events, and navigation to events domains. The first internal dataset we use contains over 170k training utterances annotated with flat representations, covering over 140 distinct intents from a variety of domains including weather, communication, music, weather, and device control. The second internal dataset contains over 67k training utterances with fully hierarchical representations, and covers over 60 intents all in the communication domain.

The second and third public datasets are SNIPS Natural Language Understanding benchmark1 (SNIPS-NLU) and the Airline Travel Information Systems (ATIS) dataset \citep{hemphill1990atis}. We follow the same procedure that was mentioned above for preparing the decoupled data for both of these datasets.

\begin{table*}[t]
   \centering
   \caption{Frame accuracy of the decoupled models on semantic parsing tasks. $\dagger$ indicates results from \citet{hakkani2016multi}; $\ddagger$, from \citet{goo2018slot}; $*$, from \citet{zhang2018joint}; $\times$, from \citet{Chen2019BERTFJ}.}
   \small
   \begin{subtable}[t]{.32\linewidth}
      \centering
      \caption{Accuracy on TOP.}\label{tbl:top}
      \begin{tabular}{lc}
      \toprule
      Model & Acc. \\
      \midrule
      RNNG & 80.86\\
      RNNG + Ensembling & 83.84 \\
      RNNG + ELMo & 83.93 \\
      \midrule
      Decoupled biLSTM & 79.51 \\
      Decoupled transformer & 64.50 \\
      Decoupled ELMo & 84.85 \\
      Decoupled RoBERTa & 84.52 \\
      Decoupled BART & \textbf{87.10} \\
      \midrule
      Best Seq2SeqPtr & 86.67 \\
      \bottomrule
      \end{tabular}
   \end{subtable}%
  \begin{subtable}[t]{.36\linewidth}
      \centering
      \caption{Accuracy on ATIS and SNIPS.}\label{tbl:top}
      \begin{tabular}{lll}
      \toprule
      Model & ATIS & SNIPS \\
      \midrule
      Joint biRNN$^\dagger$ & 80.7 & 73.2\\
      Slot gated$^\ddagger$ & 82.2 & 75.5\\
      CapsuleNLU$^*$ & 83.4 & 80.9 \\
      Joint BERT$^\times$ & 88.2 & \textbf{92.8} \\
      Joint BERT CRF$^\times$ & 88.6 & 92.6 \\
      \midrule
      Decoupled BART & \textbf{89.25} & 91.00\\
      \midrule
      Best Seq2SeqPtr & 87.12 & 87.14\\
      \bottomrule
      \end{tabular}
   \end{subtable}%
   \begin{subtable}[t]{.3\linewidth}
      \centering
      \caption{Accuracy on internal datasets.}\label{tbl:internal}
      \begin{tabular}{lc}
      \toprule
      Model & Acc. \\
      \midrule
      \multicolumn{2}{c}{Multi-domain (170k)}\\
      \midrule
      Decoupled ELMo & 86.03 \\
      Decoupled RoBERTa & {87.32}\\
      Decoupled BART &  \textbf{88.29} \\
      \midrule
      \multicolumn{2}{c}{Single-domain (67k)}\\
      \midrule
      Decoupled ELMo & 90.52 \\
      Decoupled RoBERTa & {91.51} \\
      Decoupled BART & \textbf{92.16} \\
      \bottomrule
      \end{tabular}
   \end{subtable}
\end{table*}
\begin{table*}[h!]
    \centering
    \caption{Decoupled model architecture results over the SB-TOP dataset. FA is exact match between canonicalized predicted and tree structures. Ref Only FA does not distinguish between implicit/explicit references. Intent accuracy is accuracy over top level intents while Inner Parse Accuracy is FA not considering top level intent.}
   \label{fig:sb-top}
   \small
   \begin{tabular}{@{}lcrrrr@{}}
   \toprule
   Model                   & Oracle@Beam & FA    & Ref-only FA & Intent Acc. & Inner Parse Acc. \\ \midrule
   Humans                  & 1           & 55.04 & 57.4        & 84.32           & 60.12               \\ \midrule
   Decoupled biLSTM        & 1           & 48.48 & 49.19       & 78.60           & 52.74                \\
                           & 5           & 60.24 & 69.88       & 93.71           & 72.01                \\
   Decoupled ELMo & 1           & 51.22 & 52.03       & 80.93           & 55.07                \\
                           & 5           & 62.58 & 70.08       & 94.73           & 72.11                \\
   Decoupled BART & 1           & \textbf{53.45} & \textbf{54.18}       & \textbf{82.46}           & \textbf{56.84}                \\
                           & 5           & \textbf{65.19} & \textbf{72.78}       & \textbf{96.67}           & \textbf{76.45}                \\\bottomrule
   \end{tabular}
\end{table*}

As can be seen from Table \ref{tbl:top}, our proposed approach outperforms the previous state-of-the-art results on the ATIS, comparable to state-of-the-art on SNIPS, and TOP semantic parsing task, which had been obtained with the Seq2SeqPtr model by \citet{rongali2020don}. Comparing the decoupled model to RNNGs, we note that a single decoupled model, using either biLSTMs or transformers (with RoBERTa or BART pretraining) is able to outperform the RNNG. In fact, the decoupled model even outperforms an ensemble of seven RNNGs. The decoupled biLSTM extended with ELMo inputs is able to outperform the transformer model initialised with RoBERTa pretraining. However, the best performance is achieved by using the transformer model with BART-large pretraining, with the decoupled model fine-tuned jointly on top of it \citep{lewis2019bart}. In order to understand how much of these gains are due to the semantic representation, we perform an ablation study by evaluating the biLSTM and RoBERTa-based models on TOP data using the standard logical form representation, and find a drop in frame accuracy of 0.32 and 0.55 respectively.

The TOP dataset contains to the order of 30k examples in its training set. In order to further tease out the differences between the biLSTM and transformer approaches, and to see how they compare when more training data is available, we also evaluate these models on our two larger internal datasets. Table \ref{tbl:internal} shows that the RoBERTa-based model does indeed benefit from the extra training data, being able to outperform the biLSTM-based model on the two datasets. In both cases, the decoupled model with BART pretraining achieves the top performance. 

The same procedure was used over our SB-TOP dataset, with the only variant being we concatenated SB-TOP and TOP and jointly trained over both datasets. Table~\ref{fig:sb-top} shows the test results. 
\subsection{Slot carryover}

\begin{figure}
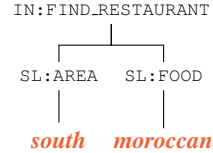
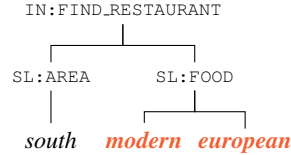

   \centering
   \begin{subfigure}{0.49\textwidth}
      \centering\small
      \tikzset{level distance=25pt,sibling distance=0pt}
      \tikzset{edge from parent/.style={draw, edge from parent path={(\tikzparentnode.south) -- +(0,-8pt) -| (\tikzchildnode)}}}
      \Tree [.\texttt{\scriptsize IN:FIND\_RESTAURANT} [.\texttt{\scriptsize SL:AREA} \emph{\textcolor{RedOrange}{\textbf{south}}} ] [.\texttt{\scriptsize SL:FOOD} \emph{\textcolor{RedOrange}{\textbf{moroccan}}} ] ]
      \caption{``i want a restaurant in the south part of town that serves moroccan food''.}\label{subfig:dstc2-1}
   \end{subfigure}
   \begin{subfigure}{0.49\textwidth}
      \centering\small
      \tikzset{level distance=25pt,sibling distance=0pt}
      \tikzset{edge from parent/.style={draw, edge from parent path={(\tikzparentnode.south) -- +(0,-8pt) -| (\tikzchildnode)}}}
      \Tree [.\texttt{\scriptsize IN:FIND\_RESTAURANT} [.\texttt{\scriptsize SL:AREA} \emph{south} ] [.\texttt{\scriptsize SL:FOOD} \emph{\textcolor{RedOrange}{\textbf{modern}}} \emph{\textcolor{RedOrange}{\textbf{european}}} ] ]
      \caption{``how about modern european''}\label{subfig:dstc2-2}
   \end{subfigure}
   \caption{Example DSTC2 session, annotated for the decoupled model.}\label{fig:dstc2}
\end{figure}
\begin{table*}[t]
   \centering
   \caption{Performance of the decoupled models on a state tracking task (DSTC2).}\label{tbl:dstc2}
   \begin{tabular}{lccccc}
   \toprule
   Model & Accuracy & \multicolumn{4}{c}{Slot distance} \\
   & & 0 & 1 & 2 & $\ge3$ \\
   \midrule
   LSTM-based \citep{naik2018} & --- & 92.42 & 91.11 & 91.34 & 87.99 \\
   Pointer network decoder \citep{alexa} & --- & 92.70 & 92.04 & 92.90 & 91.39 \\
   Transformer decoder \citep{alexa} & --- & 93.00 & 92.69 & 92.80 & 89.49 \\
   GLAD \citep{glad} & 74.5 & -- & -- & -- & --\\
   \midrule
   Decoupled biLSTM & 88.3 & {93.34} & {94.73} & {95.28} & {95.73} \\
   Decoupled RoBERTa & 89.8 & 91.98 & 92.94 & 93.58 & 94.28 \\
   Decoupled BART & \textbf{90.2} & 94.21 & 95.47 & 95.90 & 97.05 \\
   \bottomrule
   \end{tabular}
\end{table*}

To evaluate the ability of the decoupled models to work on session-based data, we evaluate them on a task which requires drawing information for multiple utterances. The DSTC2 dataset \citep{dstc2} contains a number of dialogues annotated with dialogue state -- slightly over 2k sessions in the training set. They involve users searching for restaurants, by specifying constraints such as cuisine type and price range. Given that users will often take multiple turns to specify all constraints, determining the correct dialogue state requires the model to consider all past turns too. Consider the example of the two-turn DSTC2 session shown in Figure \ref{fig:dstc2}: the \texttt{[SL:AREA south ]} slot, introduced in the first session, is said to \emph{carry over} to the second session as it still applies to the dialogue state, despite not being explicitly mentioned.\footnote{The image shows the tree form to which we converted the DSTC2 native state tracking annotations, to make them easily linearizable and thus treatable by the decoupled models.} To make previous utterances available to the model, we use a simple approach: all utterances are concatenated, with a separator token, and are fed to the encoder. 

The decoupled models are evaluated on frame accuracy and slot carryover -- the fraction of slots correctly carried over from one turn to the next. Carryover figures are split by slot distance: how many turns prior to the current one the slot under consideration first appeared. As shown in Table \ref{tbl:dstc2}, the RoBERTa decoupled model outperforms the biLSTM model on frame accuracy, while the biLSTM model takes the lead in terms of raw slot carryover performance. BART outperforms both, achieving the best overall performance.

For informative purposes, we also include results from standard dialogue state tracking models. The results show that the decoupled models, despite not being specifically designed for the task of dialogue state tracking, compare favorably to other approaches in the literature. While our models outperform them on most metrics, it should be noted that they are very different in nature: the decoupled models attend over all utterances leading up to and including the current turn, while state tracking models generally only have access to the current utterance and the previous system actions -- in the case of \citet{glad} -- or a fixed length dialogue representation. It is interesting to note that the decoupled models perform better on distant slots: this suggests that the models may be paying more attention to the beginning of the sentences, which may be an artifact of their pretraining.

\section{Related Work}

Traditional work on semantic parsing, either for the purposes of question answering or task-oriented request understanding, has focused on mapping utterances to logical form representations \citep{mooney96,zettlemoyer2005learning,kwiatkowksi2010inducing,liang2016learning,noord2018exploring}. Logical forms, while very expressive, are also complex. Highly trained annotators are required for the creation of training data, and as a result there is a lack of large scale datasets that make use of these formalisms.

Intent-slot representations such as those used for the ATIS dataset \citep{atis} or the datasets released as part of the DSTC challenges \citep{dstc2,dstc8} have less expressive power, but have the major advantage of being simple enough to enable the creation of large-scale datasets. \citet{gupta2018semantic} introduce a hierarchical intent-slot representation, and show that it is expressive enough to capture the majority of user-generated queries in two domains.

Recent approaches to semantic parsing have focused on using techniques such as RNNGs \citep{gupta2018semantic}, RNNGs augmented with ensembling and re-ranking techniques or contextual embeddings \citep{einolghozati2018improving}, sequence-to-sequence recurrent neural networks augmented with pointer mechanisms \citep{jia2016data}, capsule networks \citep{zhang2019joint}, and Transformer-based architectures \citep{rongali2020don}.

\section{Conclusions}

We started this paper by exploring the limitations of compositional intent-slot representations for semantic parsing. Due to the constraints it imposes, it cannot represent certain utterances with long-term dependencies, and it is unsuitable for semantic parsing at the session (multi-utterance) level. To overcome these limitations we propose an extension of this representation, the \emph{decoupled representation}. We propose a family of sequence-to-sequence models based on the pointer-generator architecture -- using both recurrent neural network and transformer architectures -- and show that they achieve top performance on several semantic parsing tasks. Further, to advance session-based task-oriented semantic parsing, we release to the public a new dataset of roughly 20k sessions (over 60k utterances). 

\bibliographystyle{acl_natbib}
\bibliography{emnlp2020}
\clearpage
\appendix

\section{Dataset Statistics}
\small
\begin{table}[h]
\centering
\begin{tabular}{@{}llll@{}}
\toprule
                                         & Train & Valid & Test  \\ \midrule
Size                                     & 62807 & 1004  & 1004  \\
Ref Tags                               & 2900  & 146   & 108   \\
Avg Session Length                       & 1.861 & 4.024 & 4.007 \\
Avg Utterance Length                     & 8.314 & 6.600 & 6.929 \\ \midrule
Avg Intents & 1.519 & 1.723 & 1.164 \\ \bottomrule
\end{tabular}
\caption{SBTOP Datasets Statistics}\label{table:sbtop_stats}
\end{table}

\end{document}